\documentclass[conference]{IEEEtran}
\usepackage[T1]{fontenc}
\usepackage{graphicx}
\usepackage{amsmath, amssymb}
\usepackage{booktabs}
\usepackage{url}
\usepackage{hyperref}
\usepackage{subcaption}
\usepackage{cite}
\begin{document}
    \title{UE5-Forest: A Photorealistic Synthetic Stereo Dataset for UAV Forestry Depth Estimation}
\author{\IEEEauthorblockN{Yida Lin, Bing Xue, Mengjie Zhang} \IEEEauthorblockA{\small \textit{Centre for Data Science and Artificial Intelligence} \\ \textit{Victoria University of Wellington, Wellington, New Zealand}\\ linyida\texttt{@}myvuw.ac.nz, bing.xue\texttt{@}vuw.ac.nz, mengjie.zhang\texttt{@}vuw.ac.nz}
    \and \IEEEauthorblockN{Sam Schofield, Richard Green} \IEEEauthorblockA{\small \textit{Department of Computer Science and Software Engineering} \\ \textit{University of Canterbury, Canterbury, New Zealand}\\ sam.schofield\texttt{@}canterbury.ac.nz, richard.green\texttt{@}canterbury.ac.nz}
    }
    \maketitle
    \maketitle
    \begin{abstract}
        Dense ground-truth disparity maps are practically unobtainable in forestry
        environments, where thin overlapping branches and complex canopy
        geometry defeat conventional depth sensors---a critical bottleneck for
        training supervised stereo matching networks for autonomous UAV-based
        pruning. We present \textbf{UE5-Forest}, a photorealistic synthetic stereo
        dataset built entirely in Unreal Engine~5 (UE5). One hundred and fifteen
        photogrammetry-scanned trees from the Quixel Megascans library are
        placed in virtual scenes and captured by a simulated stereo rig whose
        intrinsics---63~mm baseline, 2.8~mm focal length, 3.84~mm sensor width---replicate
        the ZED Mini camera mounted on our drone. Orbiting each tree at up to 2~m
        across three elevation bands (horizontal, $+45^{\circ}$, $-45^{\circ}$) yields
        5,520 rectified $1920\times1080$ stereo pairs with pixel-perfect disparity
        labels. We provide a statistical characterisation of the dataset---covering
        disparity distributions, scene diversity, and visual fidelity---and a
        qualitative comparison with real-world Canterbury Tree Branches imagery
        that confirms the photorealistic quality and geometric plausibility of
        the rendered data. The dataset will be publicly released to provide the community
        with a ready-to-use benchmark and training resource for stereo-based forestry
        depth estimation.
    \end{abstract}
    \section{Introduction}

    Supervised stereo matching networks depend on dense disparity ground truth, yet
    collecting such labels outside controlled settings remains an unresolved challenge.
    Forest canopies present an extreme instance of this problem: thin overlapping
    branches, dense foliage, and irregular geometry prevent reliable depth acquisition
    by either LiDAR~\cite{geiger2012kitti} or structured-light sensors~\cite{scharstein2014middlebury}.
    The resulting data vacuum directly constrains UAV-based autonomous pruning~\cite{lin2024branch,steininger2025timbervision},
    an application that demands centimetre-level depth precision at 1--2~m
    working distances for safe tool positioning around radiata pine (\textit{Pinus
    radiata})---New Zealand's dominant plantation species, underpinning an NZ\$3.6~billion
    forestry sector.

    A common workaround is to treat the output of a strong foundation model as
    pseudo-ground truth~\cite{lin2024benchmark,jiang2025defom}, but teacher-model
    errors propagate directly into every student, imposing an accuracy ceiling
    that cannot be verified in the field. Game-engine simulation offers a structurally
    different solution. Unreal Engine~5 (UE5), with its Lumen global illumination
    and Nanite micro-polygon geometry~\cite{epic2022ue5}, now produces imagery
    approaching photographic realism; at the same time, its per-pixel depth
    buffer delivers mathematically exact disparity. Pairing this rendering capability
    with the Quixel Megascans library~\cite{quixel2023megascans}---115 photogrammetry-scanned
    tree models covering a wide spectrum of species and growth forms---enables the
    synthesis of forestry-specific data that is both visually authentic and label-perfect.

    Despite these technological advances, no publicly available stereo dataset
    targets the forestry domain with dense, error-free disparity labels.
    Existing synthetic benchmarks such as Scene Flow~\cite{mayer2016large}, SYNTHIA~\cite{ros2016synthia},
    and Playing for Data~\cite{richter2016playing} focus on urban driving
    scenarios and lack the fine-grained vegetation geometry that characterises
    forest scenes. This gap motivates the creation of a dedicated resource.

    In this paper we introduce \textbf{UE5-Forest}, a synthetic stereo dataset designed
    specifically for UAV forestry depth estimation. A simulated drone-mounted
    stereo rig orbits each of the 115 trees at up to 2~m, with optical parameters---63~mm
    baseline, 2.8~mm focal length, 3.84~mm sensor width---identical to the ZED Mini~\cite{stereolabs2024zedmini}
    installed on our UAV. Three elevation rings of 16 viewpoints each ($0^{\circ}$,
    $+45^{\circ}$, $-45^{\circ}$) yield 5,520 stereo pairs at $1920\times1080$
    with pixel-perfect disparity maps.

    The principal contributions are threefold:
    \begin{itemize}
        \item A photorealistic forestry stereo dataset with pixel-accurate disparity
            labels, generated in UE5 from 115 photogrammetry-scanned tree assets
            under camera parameters precisely matched to real ZED Mini hardware.

        \item A detailed statistical characterisation of the dataset, including
            disparity range analysis, scene diversity metrics, and a qualitative
            visual comparison with real-world Canterbury Tree Branches imagery
            that confirms the photorealistic quality of the rendered data.

        \item A publicly released, ready-to-use benchmark and training resource
            that fills the data vacuum for supervised stereo matching research
            in the forestry domain.
    \end{itemize}

    \section{Related Work}

    \subsection{UAV-Based Depth Estimation in Forestry}

    Stereo cameras have served as primary depth sensors on UAV platforms across
    diverse outdoor tasks, from visual navigation~\cite{fraundorfer2012vision} and
    3D terrain reconstruction~\cite{nex2014uav} to high-speed obstacle avoidance~\cite{barry2015pushbroom}.
    Within forestry, TIMBERVision~\cite{steininger2025timbervision} deployed stereo
    depth for stand-level canopy monitoring, while integrated branch detection~\cite{lin2024branch,lin2025yolosgbm}
    and segmentation~\cite{lin2025segmentation} pipelines couple disparity
    output with object recognition for pruning automation. A recent benchmark~\cite{lin2024benchmark}
    evaluated seven stereo methods on vegetation imagery but was bound to pseudo-ground
    truth from the DEFOM foundation model~\cite{jiang2025defom}, underscoring
    the persistent absence of reliable reference data in this domain.

    \subsection{Stereo Matching and the Need for Dense Labels}

    Stereo matching architectures---from cost-volume methods~\cite{kendall2017gcnet,chang2018psmnet,guo2019gwcnet}
    to iterative refinement designs such as RAFT-Stereo~\cite{lipson2021raft} and
    IGEV-Stereo~\cite{xu2023igev}---have achieved impressive accuracy on established
    benchmarks. Lightweight variants~\cite{wang2019anynet,duggal2019deeppruner}
    and attention-based approaches~\cite{chen2024mocha} extend the design space
    further. Yet every supervised architecture depends on dense disparity labels
    for training---labels that remain scarce for natural vegetation scenes. This
    data bottleneck, rather than architectural limitations, is the primary
    obstacle to deploying stereo networks in forestry.

    \subsection{Synthetic Datasets for Depth Estimation}

    Synthetic data has proven indispensable for stereo research. The Scene Flow
    dataset~\cite{mayer2016large}---rendered urban scenes with exact disparity---remains
    the standard pre-training corpus. Large-scale game-engine datasets such as SYNTHIA~\cite{ros2016synthia}
    and Playing for Data~\cite{richter2016playing} demonstrated that photorealistic
    renders transfer well to autonomous-driving perception. Domain randomisation~\cite{tobin2017domain}
    and domain-invariant feature learning~\cite{zhang2020domaininvariant}
    further narrow the sim-to-real gap.

    Two recent developments make this strategy particularly attractive for
    forestry. First, UE5's Nanite and Lumen systems~\cite{epic2022ue5} render
    high-polygon vegetation with physically based lighting at interactive rates.
    Second, the Quixel Megascans library~\cite{quixel2023megascans} supplies
    photogrammetry-scanned tree assets that preserve authentic bark textures, leaf
    geometry, and branch topology. However, no prior work has harnessed these
    resources to construct a stereo dataset tailored to forestry. Table~\ref{tab:dataset_comparison}
    positions UE5-Forest relative to existing synthetic stereo benchmarks.

    \begin{table}[htbp]
        \caption{Comparison of UE5-Forest with existing synthetic stereo
        datasets.}
        \label{tab:dataset_comparison}
        \centering
        \footnotesize
        \begin{tabular}{lcccl}
            \toprule \textbf{Dataset}                  & \textbf{Pairs} & \textbf{Resolution}       & \textbf{Domain}   & \textbf{Labels}    \\
            \midrule Scene Flow~\cite{mayer2016large}  & 39,049         & $960\times540$            & Urban             & Disparity, flow    \\
            SYNTHIA~\cite{ros2016synthia}              & 213,400        & $1280\times760$           & Urban             & Depth, seg.        \\
            Playing for Data~\cite{richter2016playing} & 24,966         & $1914\times1052$          & Urban             & Seg.               \\
            \textbf{UE5-Forest (ours)}                 & \textbf{5,520} & $\mathbf{1920\times1080}$ & \textbf{Forestry} & \textbf{Disparity} \\
            \bottomrule
        \end{tabular}
    \end{table}

    \section{Dataset Construction}

    \subsection{Overview}

\begin{figure}[htbp]
        \centering
        \includegraphics[width=1\columnwidth]{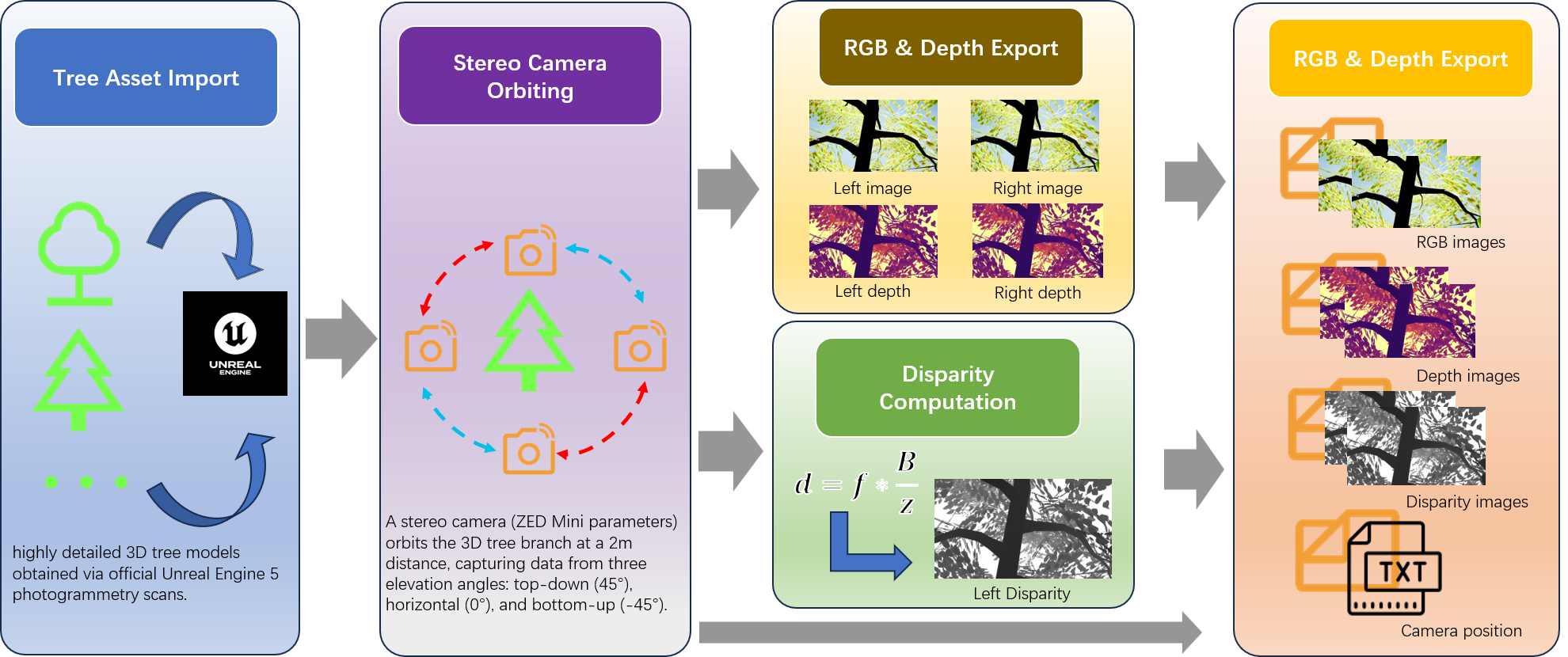}
        \caption{Overview of the UE5-Forest dataset generation pipeline.}
        \label{fig:pipeline}
    \end{figure}

    The dataset generation pipeline, illustrated in Fig.~\ref{fig:pipeline}, proceeds
    in three stages: (1) photogrammetry-scanned tree assets are imported into
    UE5; (2) a simulated stereo camera orbits each tree under controlled lighting;
    and (3) RGB image pairs together with per-pixel depth maps are exported and
    converted into disparity ground truth.

    \subsection{Stereo Geometry and Camera Model}

    Given a rectified stereo pair
    $I_{L}, I_{R}\in \mathbb{R}^{H\times W\times 3}$, stereo matching estimates a
    disparity map $D\in\mathbb{R}^{H\times W}$ where each pixel $(x,y)$ in the left
    image corresponds to $(x-D(x,y),\,y)$ in the right image. Depth is recovered
    via triangulation:
    \begin{equation}
        Z(x,y) = \frac{f \cdot B}{D(x,y)}
    \end{equation}
    where $f$ is the focal length in pixels and $B$ is the stereo baseline in metres.
    The simulated camera replicates the ZED Mini specifications (Table~\ref{tab:camera_params}).
    The focal length in pixels is:
    \begin{equation}
        f_{\text{px}}= \frac{f_{\text{mm}}\times W_{\text{px}}}{w_{\text{sensor}}}
        = \frac{2.8 \times 1920}{3.84}= 1400 \;\text{px}
    \end{equation}
    At the maximum operating distance of 2~m, the minimum expected disparity is:
    \begin{equation}
        D_{\text{min}}= \frac{1400 \times 0.063}{2.0}= 44.1 \;\text{px}
    \end{equation}

    \begin{table}[htbp]
        \caption{Stereo camera parameters. The UE5 simulation exactly matches the
        real ZED Mini to minimise the optical domain gap.}
        \label{tab:camera_params}
        \centering
        \small
        \begin{tabular}{lcc}
            \toprule \textbf{Parameter} & \textbf{ZED Mini} & \textbf{UE5 Simulation} \\
            \midrule Baseline           & 63~mm             & 63~mm                   \\
            Focal length                & 2.8~mm            & 2.8~mm                  \\
            Sensor width                & 3.84~mm           & 3.84~mm                 \\
            Resolution                  & $1920\times1080$  & $1920\times1080$        \\
            Focal length (px)           & 1400              & 1400                    \\
            \bottomrule
        \end{tabular}
    \end{table}

    \subsection{Tree Assets}

\begin{figure}[htbp]
        \centering
        \includegraphics[width=1\columnwidth]{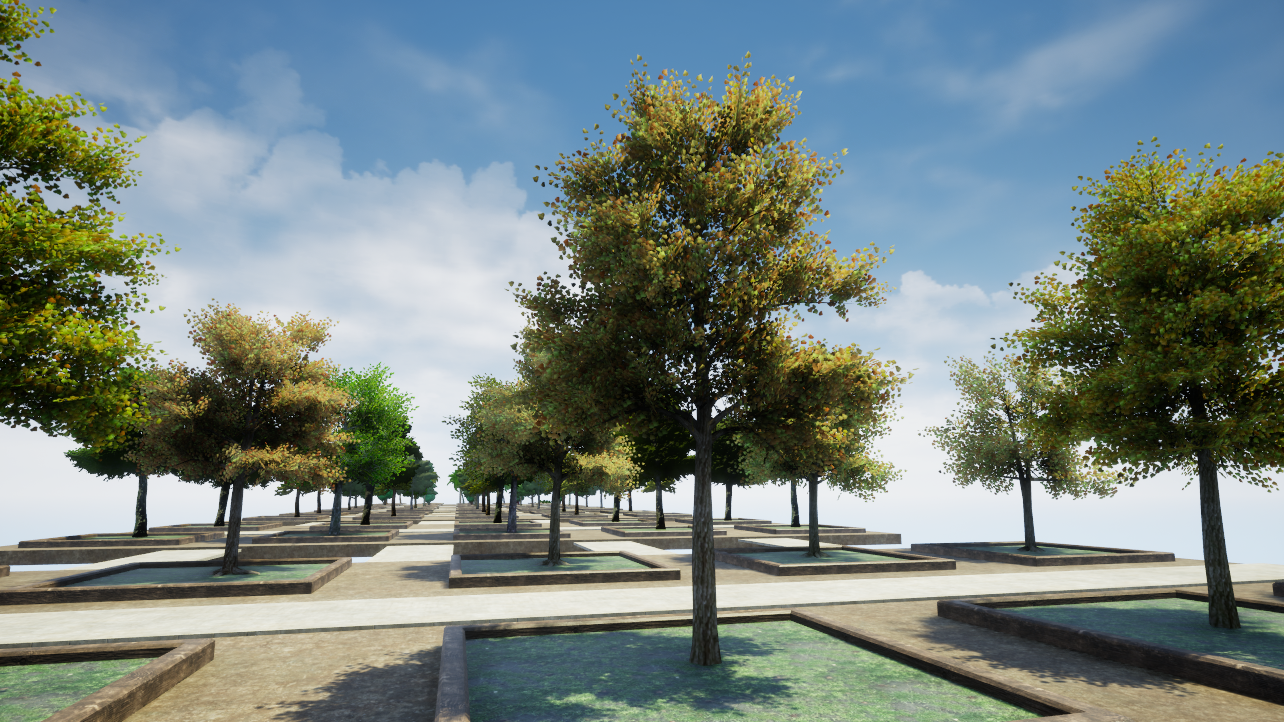}
        \caption{trees in UE5}
        \label{fig:tree_assets}
    \end{figure}

    We source 115 photogrammetry-scanned tree models from the Quixel Megascans library~\cite{quixel2023megascans},
    which provides game-ready assets created from high-resolution real-world
    scans. These trees span a diverse range of species, morphologies, and canopy
    structures, including deciduous and coniferous varieties with varying branch
    densities and foliage patterns (Fig.~\ref{fig:tree_assets}). Each asset
    retains photorealistic bark textures, leaf geometry, and branch topology
    from the original scan.

    \subsection{Drone Camera Simulation}

\begin{figure}[htbp]
        \centering
        \setlength{\tabcolsep}{1pt}
        \begin{tabular}{cccccc}
            $\vcenter{\hbox{\includegraphics[width=0.19\columnwidth]{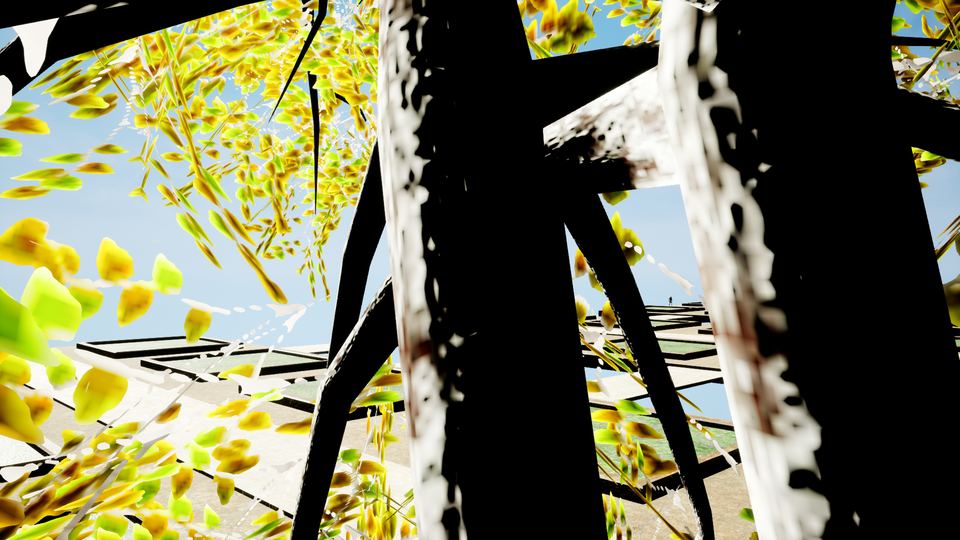}}}$       & $\vcenter{\hbox{\includegraphics[width=0.19\columnwidth]{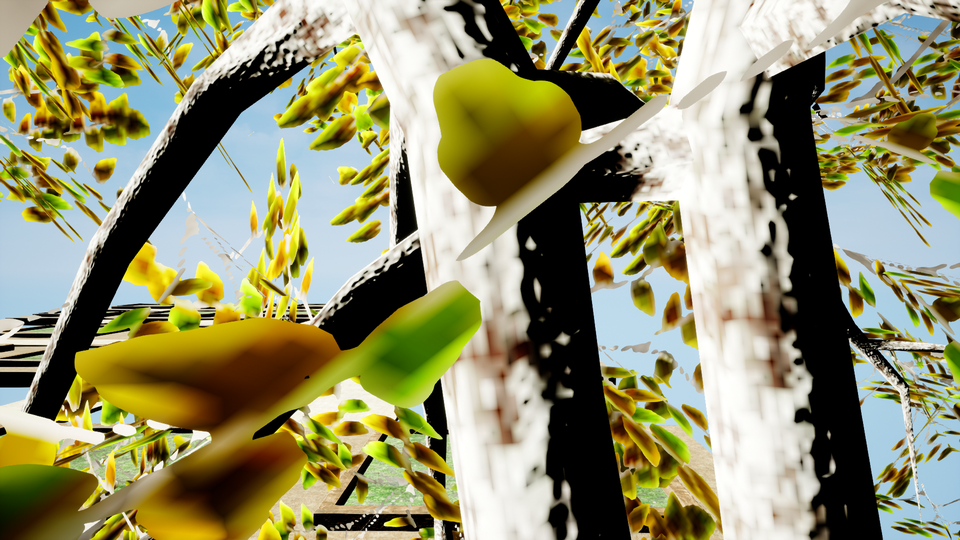}}}$ & $\vcenter{\hbox{\includegraphics[width=0.19\columnwidth]{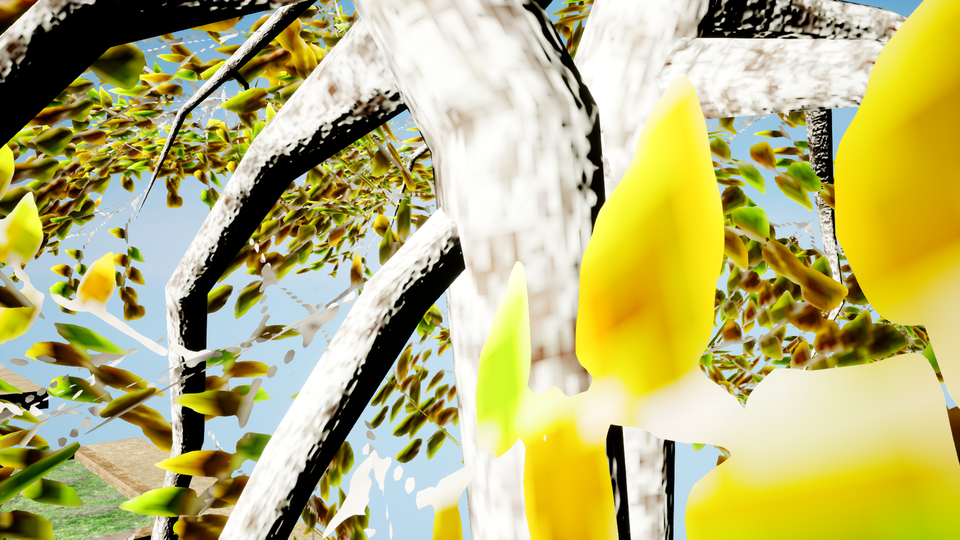}}}$ & $\vcenter{\hbox{\includegraphics[width=0.19\columnwidth]{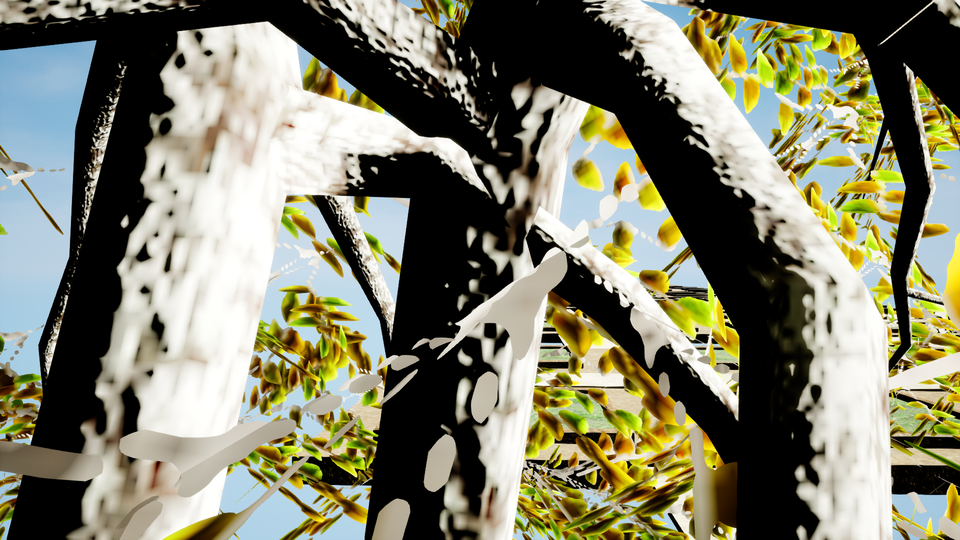}}}$ & $\cdots$ & $\vcenter{\hbox{\includegraphics[width=0.19\columnwidth]{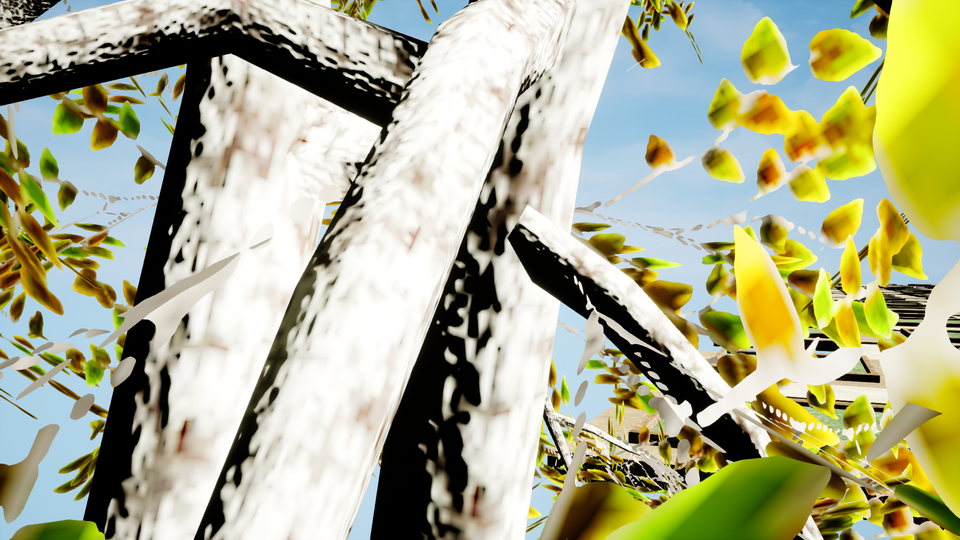}}}$ \\
            [4ex] $\vcenter{\hbox{\includegraphics[width=0.19\columnwidth]{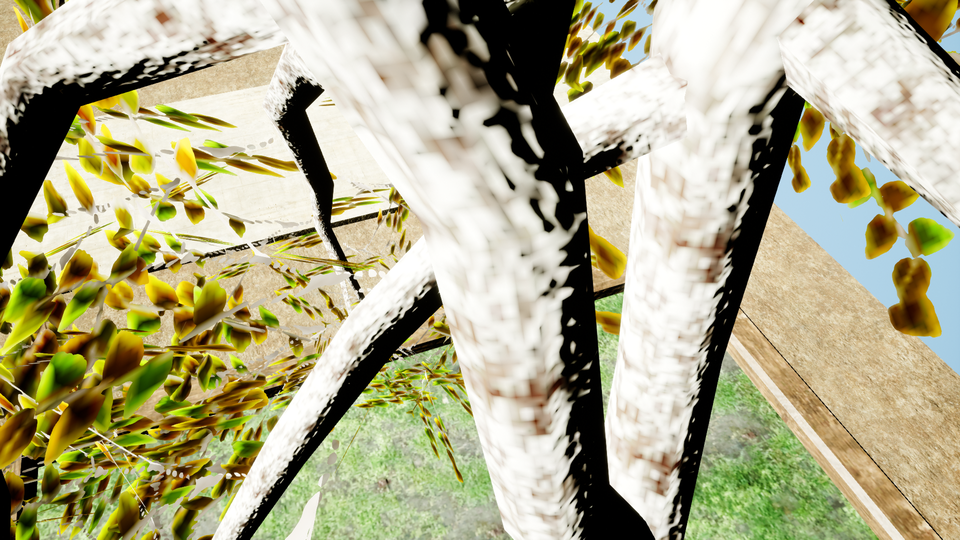}}}$ & $\vcenter{\hbox{\includegraphics[width=0.19\columnwidth]{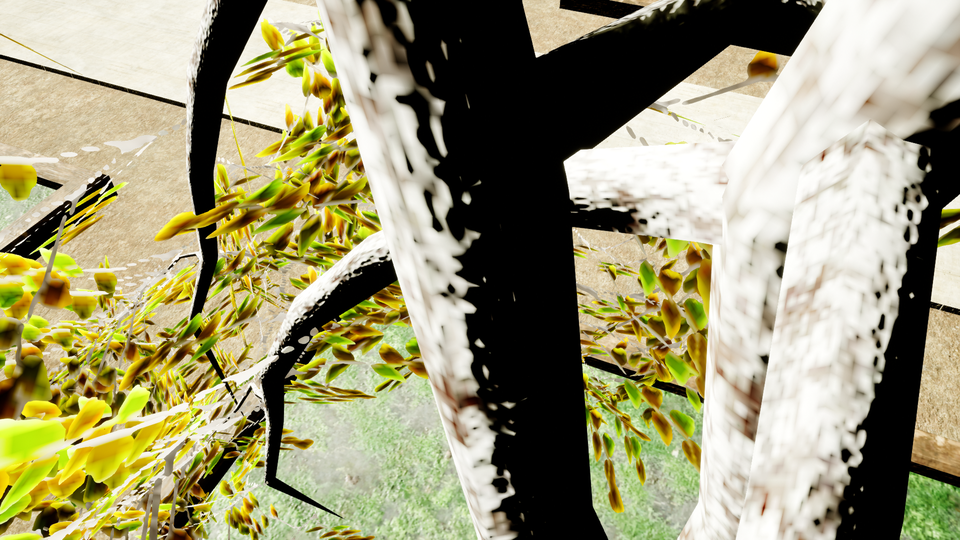}}}$ & $\vcenter{\hbox{\includegraphics[width=0.19\columnwidth]{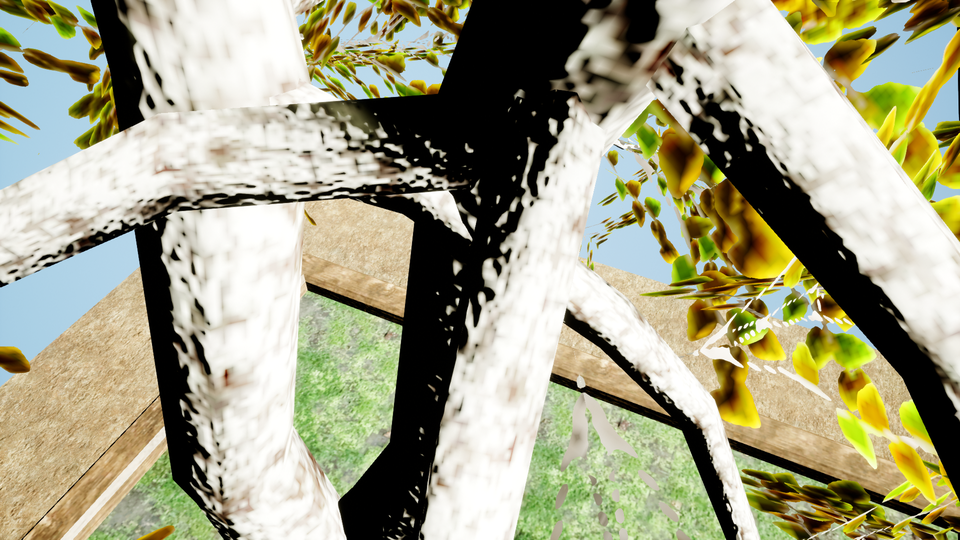}}}$ & $\vcenter{\hbox{\includegraphics[width=0.19\columnwidth]{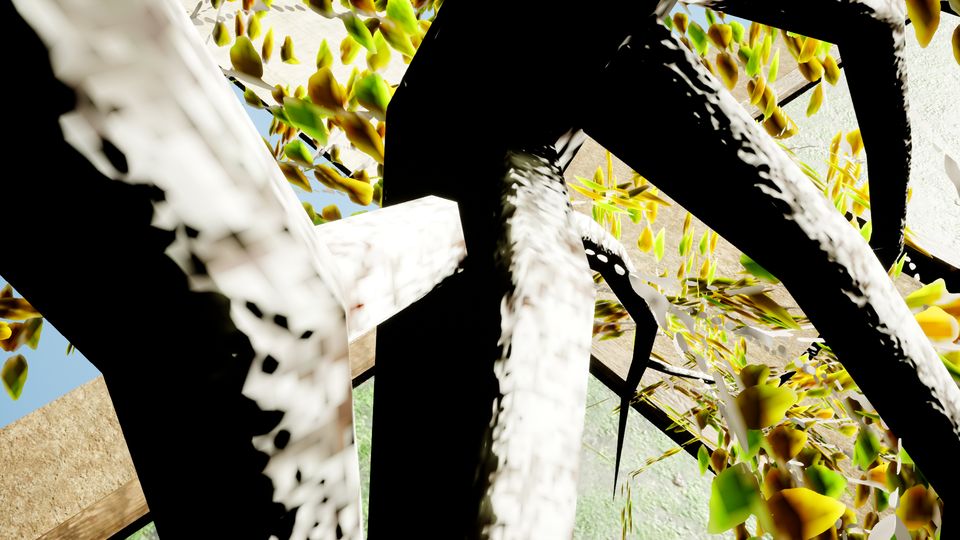}}}$ & $\cdots$ & $\vcenter{\hbox{\includegraphics[width=0.19\columnwidth]{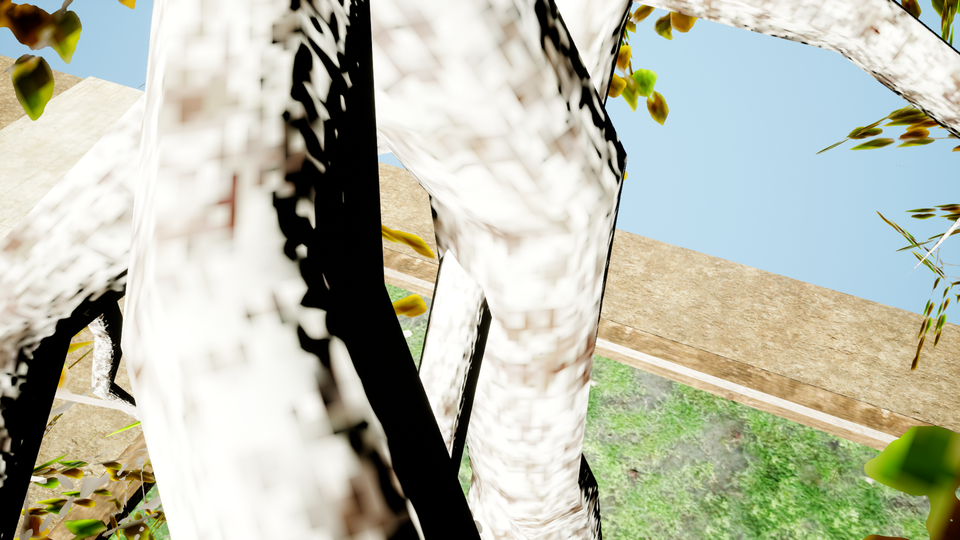}}}$ \\
            [4ex] $\vcenter{\hbox{\includegraphics[width=0.19\columnwidth]{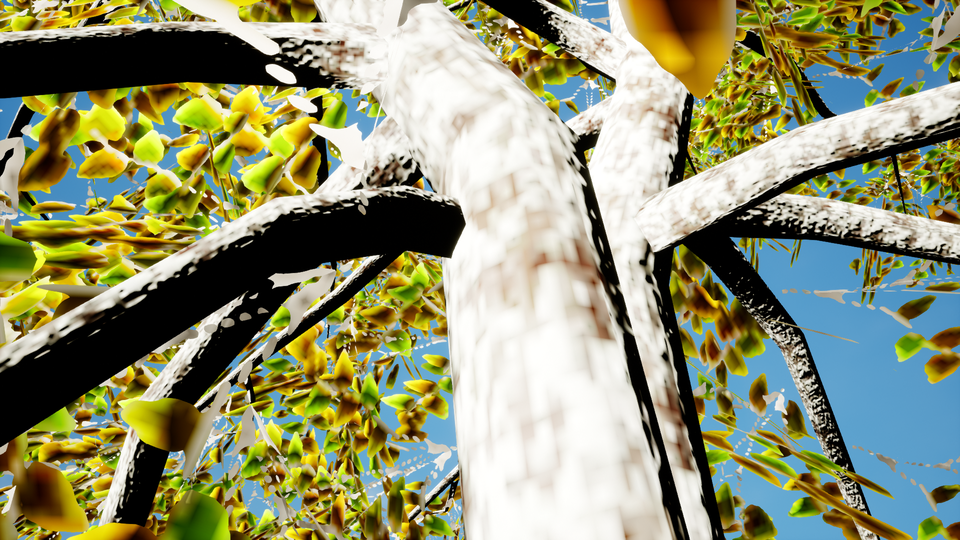}}}$ & $\vcenter{\hbox{\includegraphics[width=0.19\columnwidth]{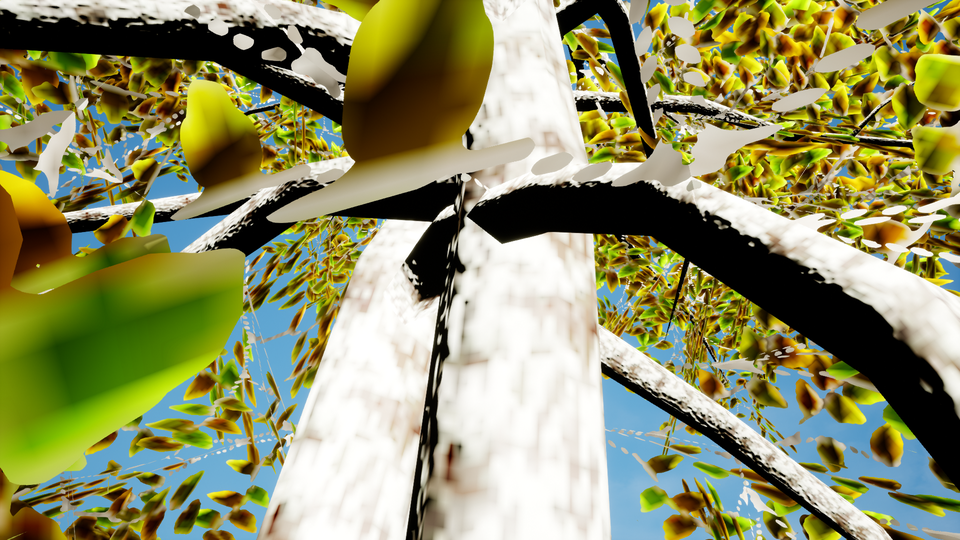}}}$ & $\vcenter{\hbox{\includegraphics[width=0.19\columnwidth]{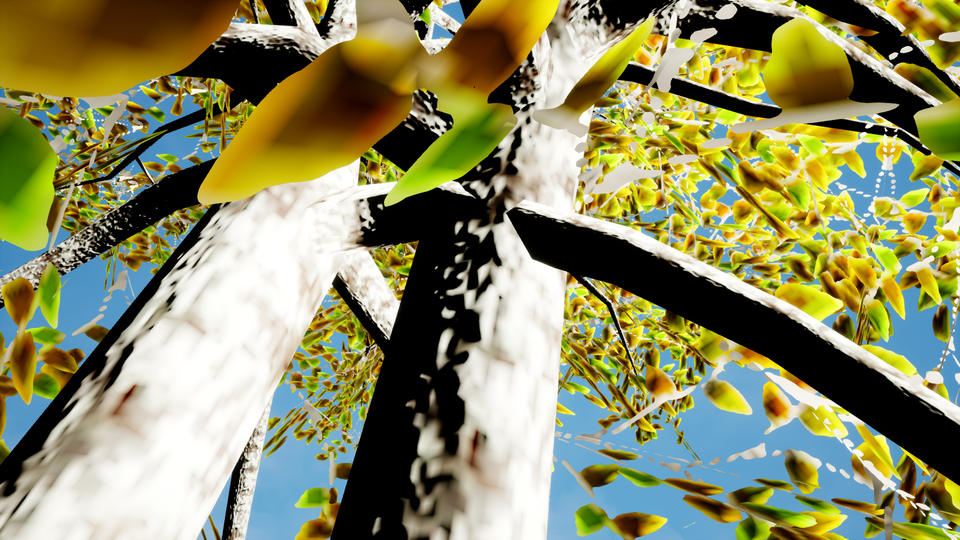}}}$ & $\vcenter{\hbox{\includegraphics[width=0.19\columnwidth]{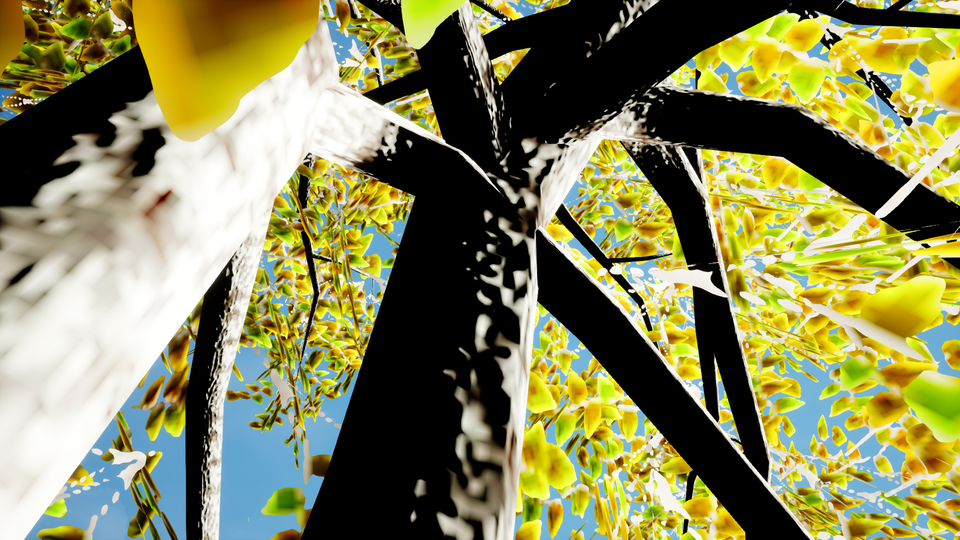}}}$ & $\cdots$ & $\vcenter{\hbox{\includegraphics[width=0.19\columnwidth]{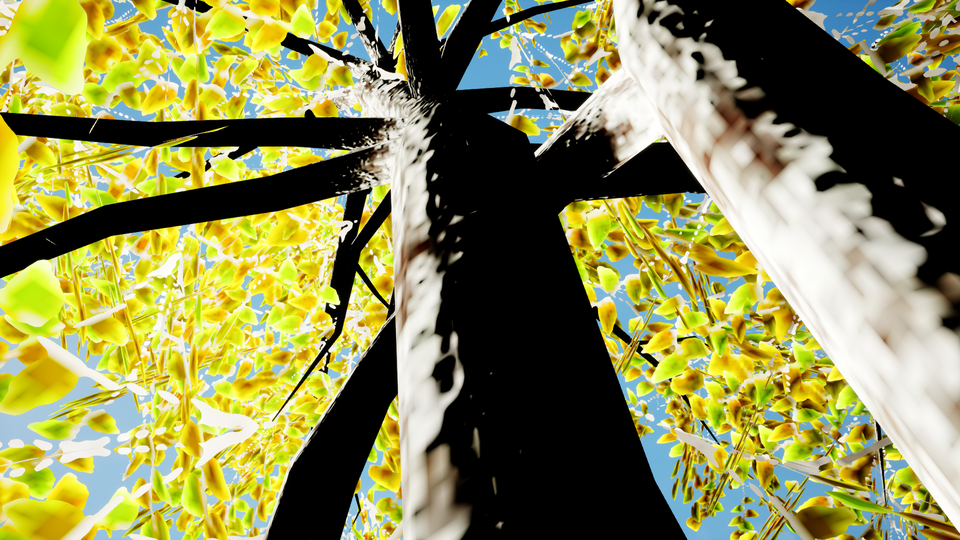}}}$ \\
        \end{tabular}
        \caption{Simulated stereo camera trajectory. Three rings of 16 viewpoints
        each at $0^{\circ}$ (H), $+45^{\circ}$ (U), and $-45^{\circ}$ (D) elevation.}
        \label{fig:camera_trajectory}
    \end{figure}

    For each tree, a simulated drone-mounted stereo camera orbits at a fixed radius
    of up to 2~m from the trunk centre, matching the operational distance range
    of our real UAV pruning system. At each elevation angle, 16 equally spaced
    viewpoints are captured ($22.5^{\circ}$ azimuth intervals), yielding three rings
    of views:
    \begin{itemize}
        \item \textbf{Horizontal ring}: 16 views at $0^{\circ}$ elevation

        \item \textbf{Elevated ring}: 16 views at $+45^{\circ}$ (looking upward)

        \item \textbf{Depressed ring}: 16 views at $-45^{\circ}$ (looking
            downward)
    \end{itemize}
    This produces 48 stereo pairs per tree, totalling $115 \times 48 = 5{,}520$ stereo
    pairs. The camera is always oriented toward the tree centre, ensuring the target
    branch structure fills the field of view (Fig.~\ref{fig:camera_trajectory}).

    \subsection{Rendering and Ground-Truth Extraction}

    UE5's Lumen global illumination and Nanite virtualised geometry systems
    provide physically based rendering with realistic lighting, shadows, and
    material interactions~\cite{epic2022ue5}. For each viewpoint, we render:
    \begin{itemize}
        \item Left and right RGB images ($1920\times 1080$, 8-bit)

        \item Per-pixel depth maps for both views (32-bit float, in metres)
    \end{itemize}
    Disparity ground truth is computed from the depth maps:
    \begin{equation}
        D_{\text{GT}}(x,y) = \frac{f_{\text{px}}\cdot B}{Z(x,y)}
    \end{equation}
    where $Z(x,y)$ is the rendered depth at pixel $(x,y)$. Pixels at infinite depth
    (sky/background) are assigned zero disparity and flagged with a validity mask.

    \subsection{Dataset Splits and Statistics}

    Table~\ref{tab:dataset_stats} summarises the dataset. We split by tree identity
    (not by image) to prevent data leakage across views of the same tree,
    following standard practice in dataset design.

    \begin{table}[htbp]
        \caption{UE5-Forest dataset statistics.}
        \label{tab:dataset_stats}
        \centering
        \small
        \begin{tabular}{lccc}
            \toprule \textbf{Split} & \textbf{Trees} & \textbf{Pairs} & \textbf{Percentage} \\
            \midrule Training       & 95             & 4,560          & 82.6\%              \\
            Validation              & 10             & 480            & 8.7\%               \\
            Test                    & 10             & 480            & 8.7\%               \\
            \midrule Total          & 115            & 5,520          & 100\%               \\
            \bottomrule
        \end{tabular}
    \end{table}

    \subsection{Data Format and Organisation}

    Each sample in UE5-Forest consists of four files: a left RGB image, a right
    RGB image, a 32-bit disparity map (stored as a \texttt{.pfm} file), and a binary
    validity mask that flags sky/background pixels. The directory structure is
    organised by tree identity and viewpoint index, facilitating both per-tree and
    per-viewpoint access. A companion metadata file records the tree asset name,
    elevation angle, azimuth angle, and all camera parameters for every sample.

    \section{Dataset Analysis}

    \subsection{Disparity Distribution}

\begin{figure}[htbp]
        \centering
        \includegraphics[width=1\columnwidth]{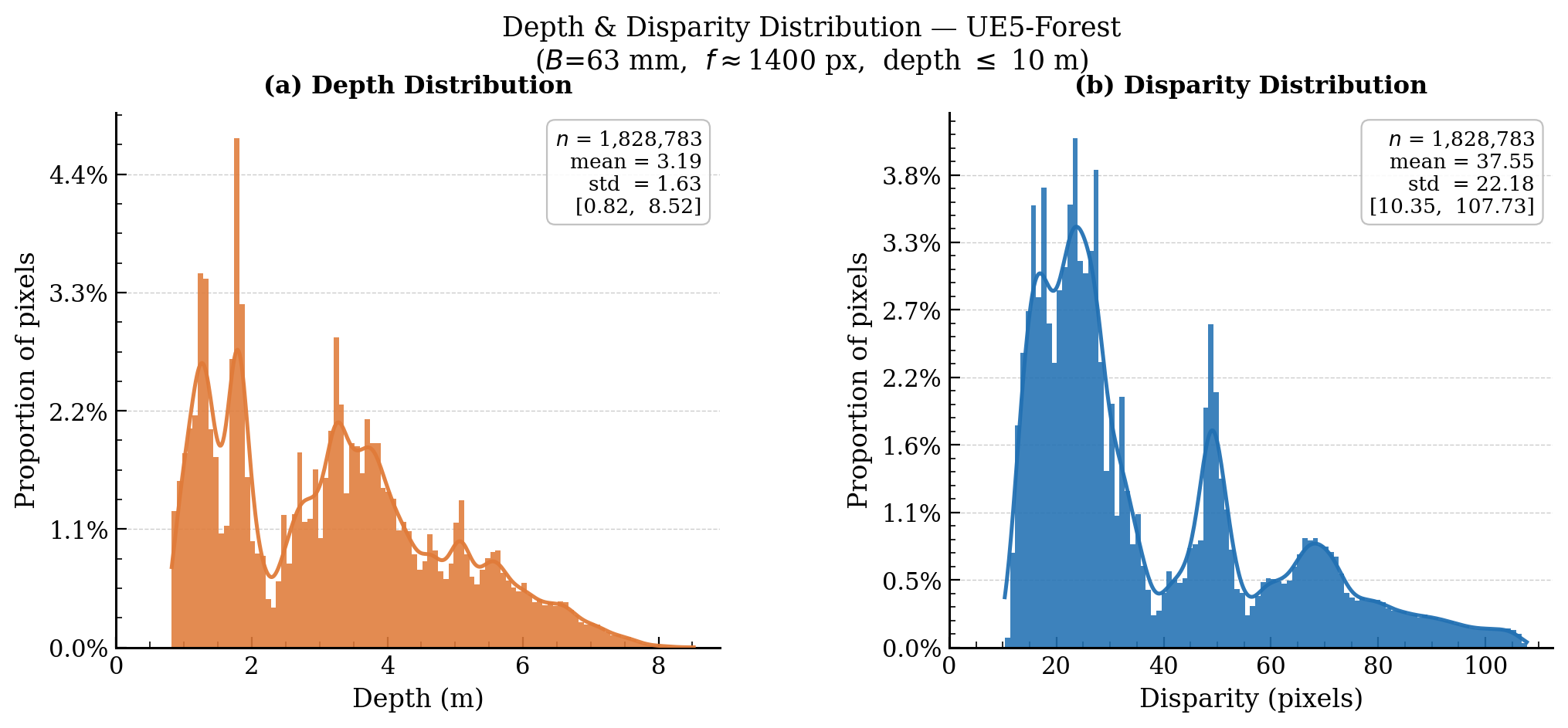}
        \caption{Ground-truth disparity distribution of the UE5-Forest dataset. The
        range is consistent with the 0.1--2~m operating distance of the
        simulated drone.}
        \label{fig:disparity_dist}
    \end{figure}

    Fig.~\ref{fig:disparity_dist} shows the ground-truth disparity distribution
    across the full dataset. The distribution spans approximately 44--700~px, consistent
    with the 0.1--2~m depth range of the simulated orbits. A concentration of values
    between 60 and 200~px corresponds to trunk and major branch surfaces at typical
    working distances, while the long tail toward higher disparities reflects
    close-range foliage and fine branches.

    \subsection{Scene Diversity}

    The 115 tree assets encompass a broad morphological range: deciduous broadleaf
    trees with dense canopies, coniferous species with needle-like foliage, and open-canopy
    specimens with prominent bare branches. This diversity ensures that models
    trained on UE5-Forest encounter varied branch densities, leaf shapes, and
    occlusion patterns. The three-ring viewpoint strategy further multiplies scene
    variety by presenting each tree from horizontal, elevated, and depressed perspectives---viewing
    geometries that correspond to different UAV flight altitudes relative to the
    canopy.

    \subsection{Visual Fidelity: Synthetic vs.\ Real Imagery}

\begin{figure}[htbp]
        \centering
        \begin{subfigure}
            {0.49\columnwidth}
            \centering
            \includegraphics[width=\linewidth]{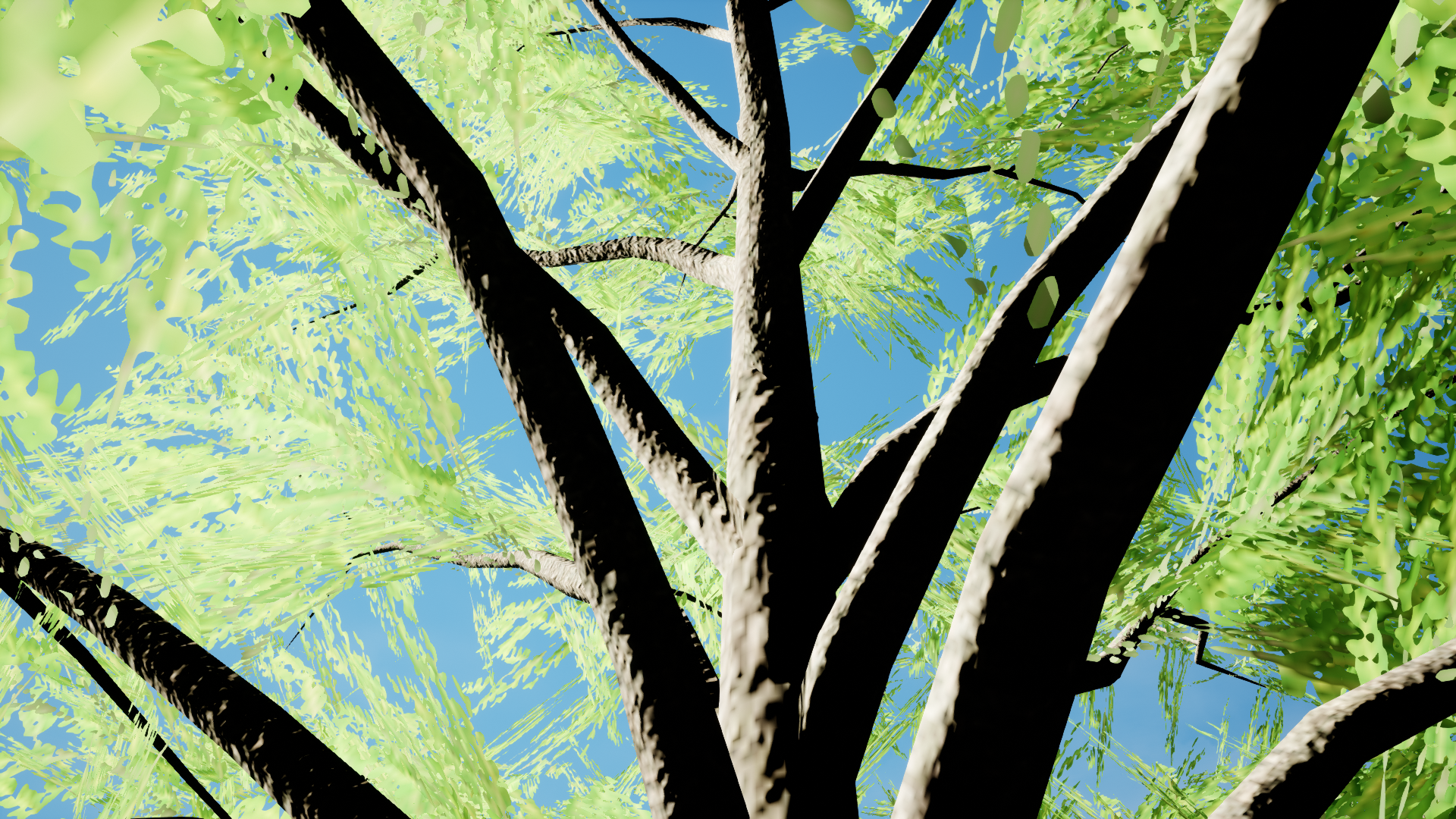}
            \caption{UE5-Forest synthetic image.}
        \end{subfigure}\hfill
        \begin{subfigure}
            {0.49\columnwidth}
            \centering
            \includegraphics[width=\linewidth]{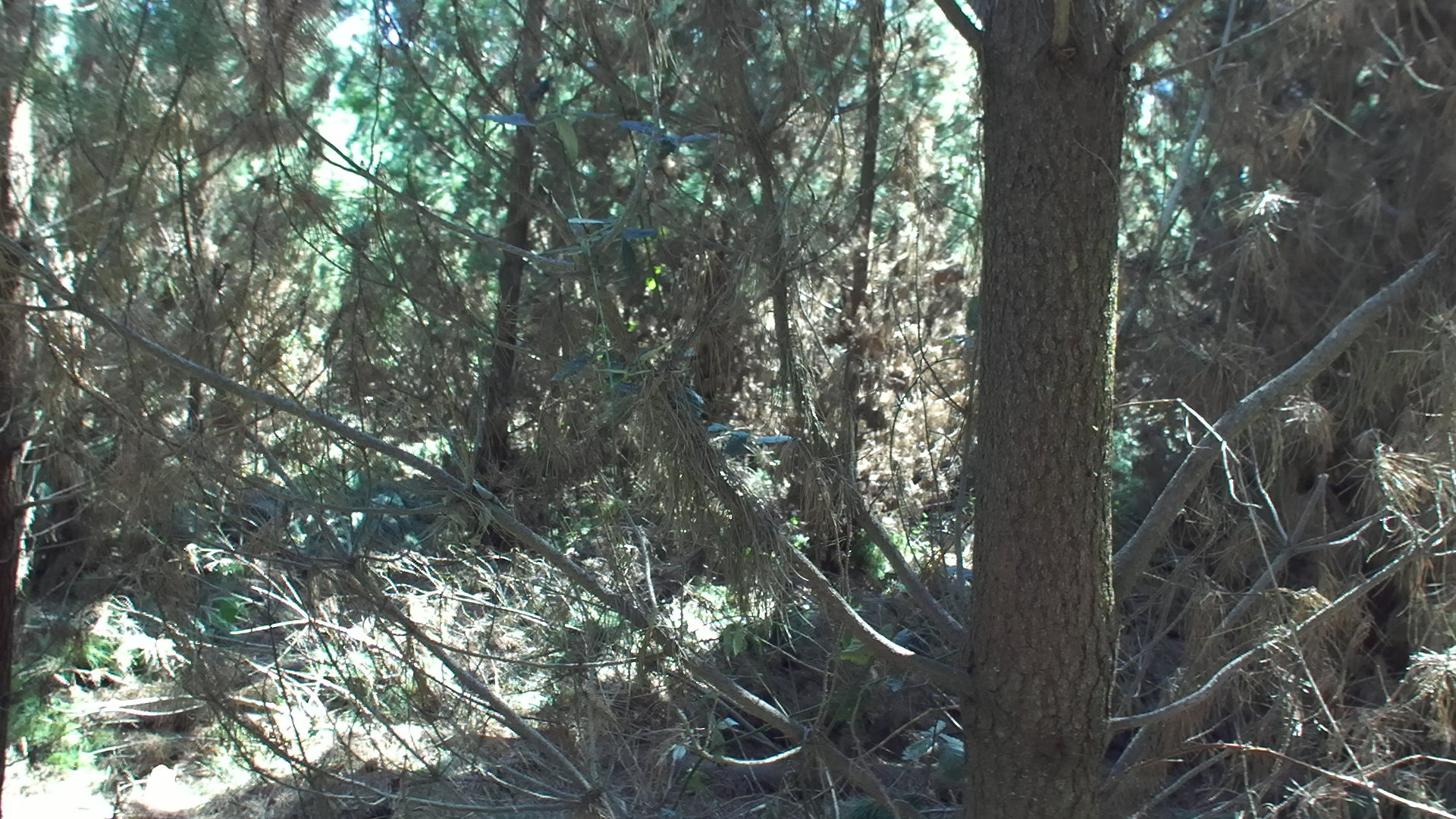}
            \caption{Canterbury real-world image.}
        \end{subfigure}
        \caption{Qualitative comparison between UE5-Forest synthetic renders (left)
        and real-world Canterbury Tree Branches images (right). Camera parameters
        are matched, producing similar perspective geometry and disparity ranges.}
        \label{fig:qualitative_comparison}
    \end{figure}

    To assess whether the synthetic data is visually plausible, we provide a side-by-side
    comparison between UE5-Forest renders and real-world images from the Canterbury
    Tree Branches dataset (Fig.~\ref{fig:qualitative_comparison}). The Canterbury
    dataset comprises 500 curated stereo pairs recorded at $1920\times1080$ with
    a ZED Mini camera (63~mm baseline) mounted on a UAV flying around radiata pine
    stands in Canterbury, New Zealand, between March and October 2024. These pairs
    were selected through multi-stage screening for motion blur, exposure
    consistency, and rectification accuracy.

    The visual comparison reveals that UE5's Lumen lighting produces
    naturalistic shadow patterns and ambient occlusion around branches, while the
    photogrammetry-scanned bark and leaf textures closely resemble their real counterparts.
    The matched camera intrinsics ensure that perspective foreshortening and
    depth-of-field characteristics are consistent between the two domains.

    \subsection{Sample Renders and Ground Truth}

\begin{figure}[htbp]
        \centering
        \begin{subfigure}
            {0.49\columnwidth}
            \centering
            \includegraphics[width=\linewidth]{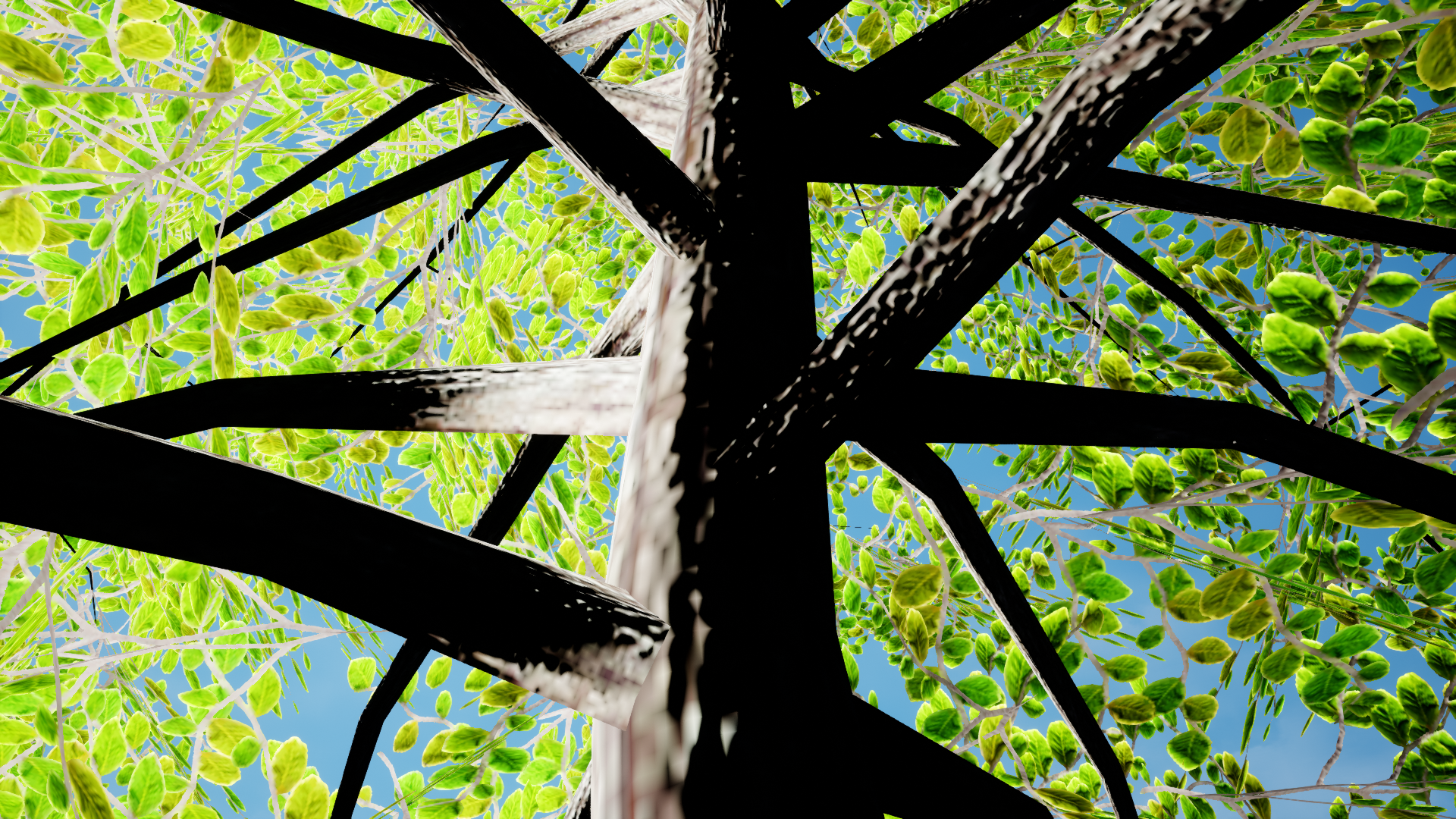}
            \caption{Left image}
        \end{subfigure}\hfill
        \begin{subfigure}
            {0.49\columnwidth}
            \centering
            \includegraphics[width=\linewidth]{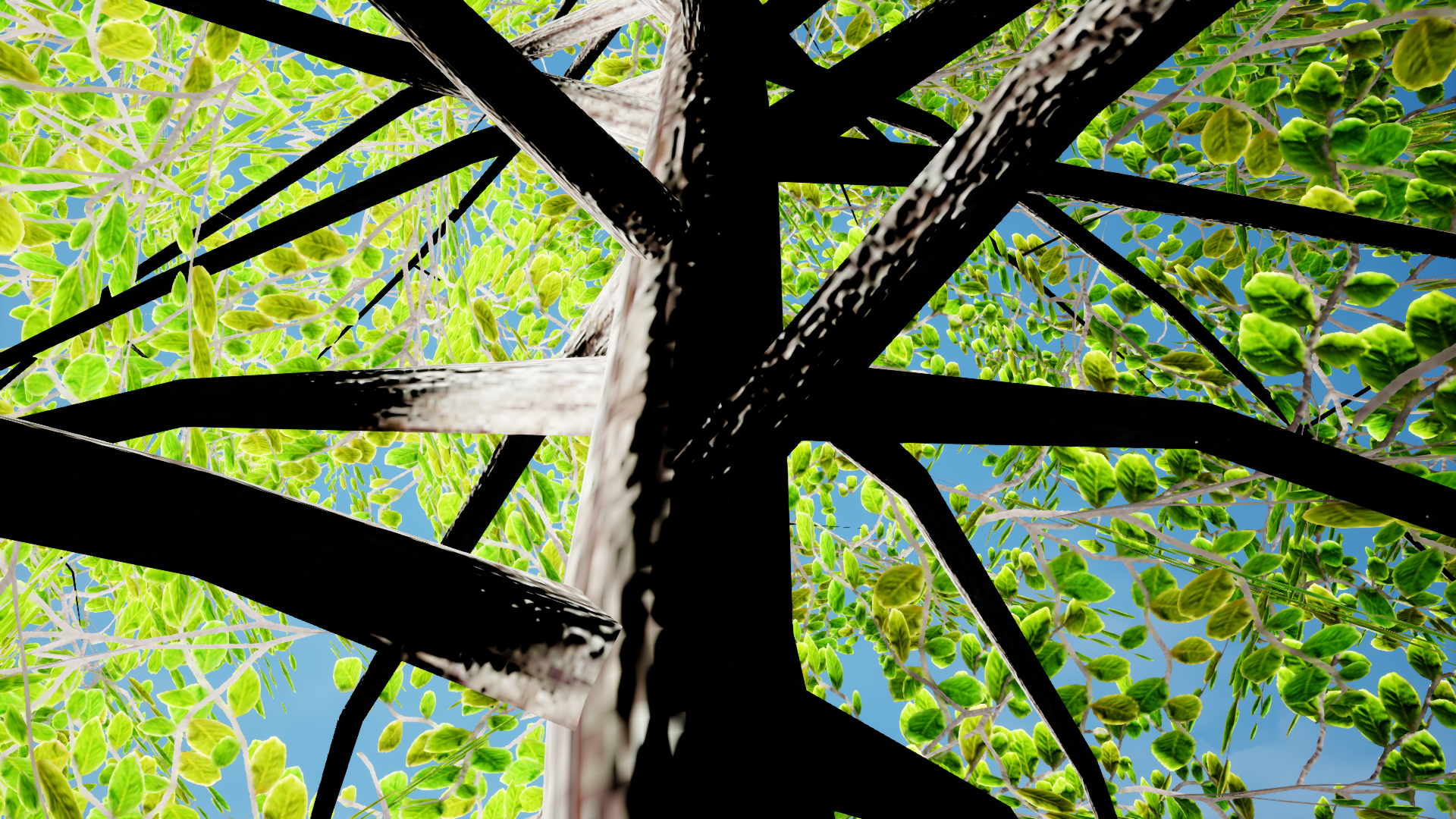}
            \caption{Right image}
        \end{subfigure}

        \vspace{0.4em}

        \begin{subfigure}
            {0.49\columnwidth}
            \centering
            \includegraphics[width=\linewidth]{
                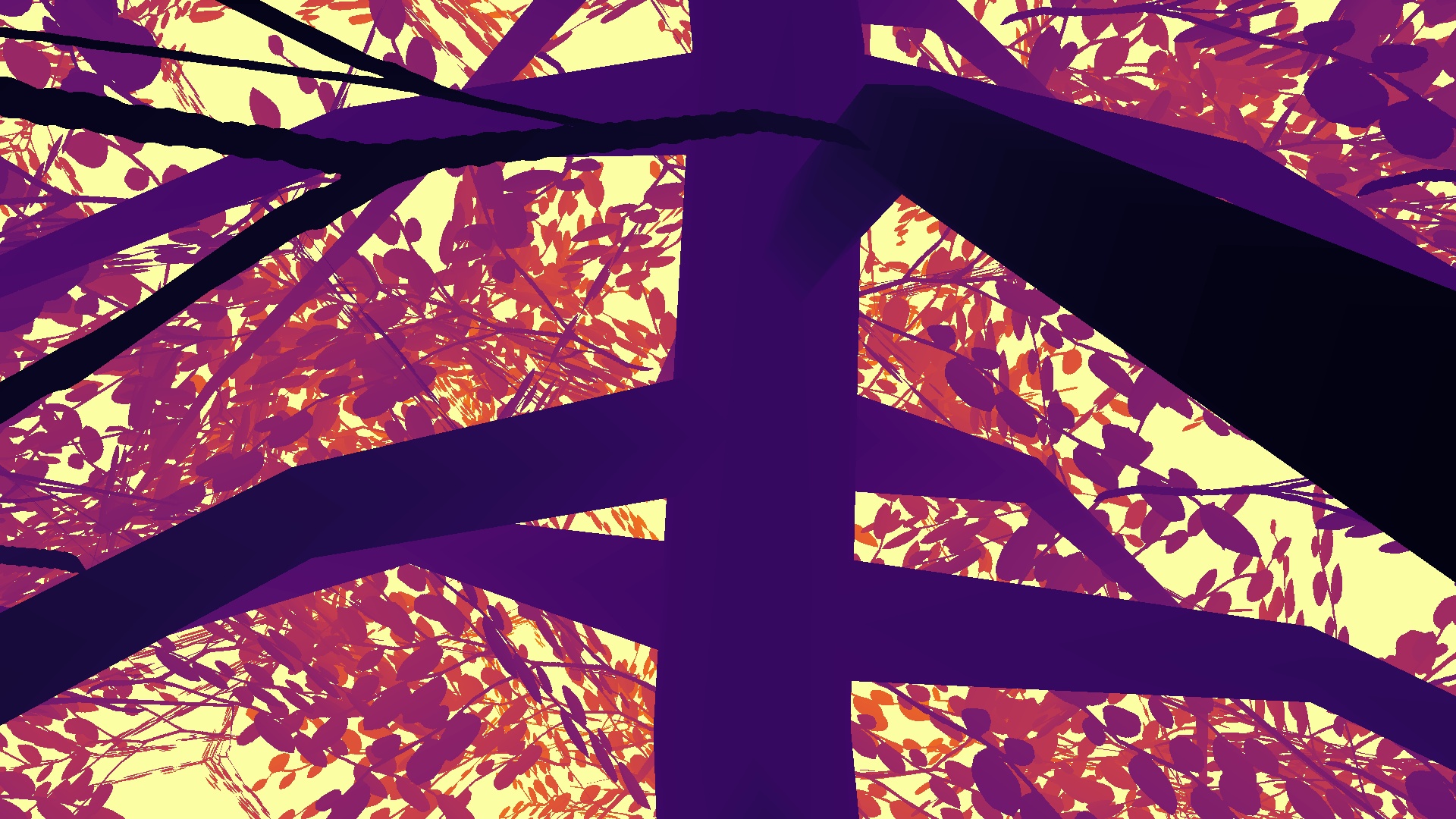
            }
            \caption{Colored depth map}
        \end{subfigure}\hfill
        \begin{subfigure}
            {0.49\columnwidth}
            \centering
            \includegraphics[width=\linewidth]{
                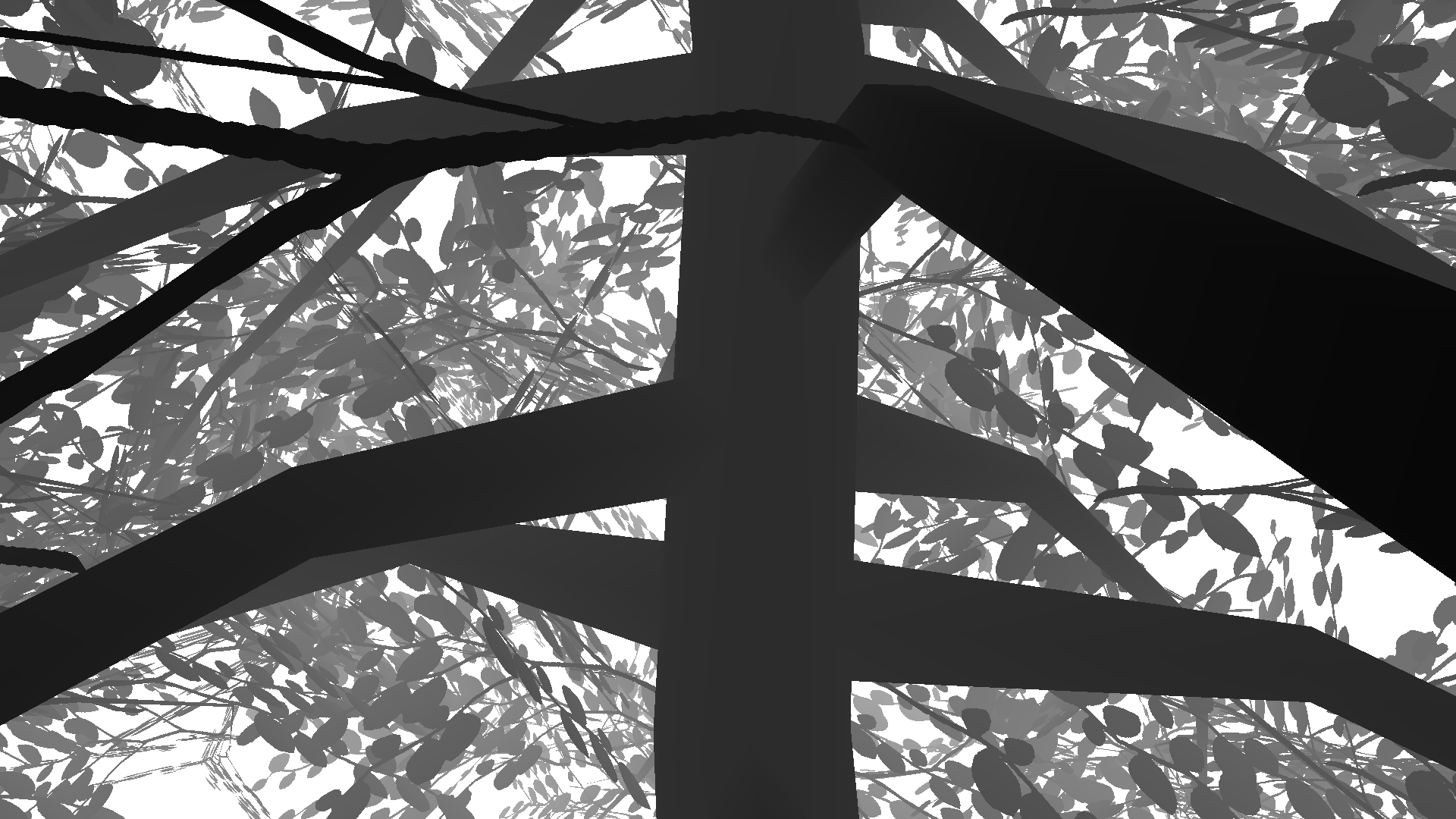
            }
            \caption{Uncolored disparity map}
        \end{subfigure}
        \caption{Sample stereo pairs and ground-truth disparity maps from UE5-Forest.
        Disparity labels are pixel-perfect, with invalid (sky) regions masked.}
        \label{fig:synthetic_examples}
    \end{figure}

    Fig.~\ref{fig:synthetic_examples} presents representative samples from the
    dataset, showing left RGB images alongside their corresponding disparity ground-truth
    maps. Fine branch structures, leaf clusters, and depth discontinuities at
    object boundaries are faithfully captured in the pixel-perfect labels.

    \section{Discussion}

    \subsection{Design Rationale}

    Three design decisions underpin the dataset's utility. The most
    consequential is \emph{asset authenticity}: because the Quixel Megascans
    trees originate from photogrammetry scans of real specimens, they carry the same
    irregular branching angles, layered foliage, and heterogeneous bark textures
    that stereo networks must interpret in the field---properties that procedurally
    generated vegetation typically lacks. Equally important is \emph{optical
    alignment}: reproducing the ZED Mini's baseline (63~mm), focal length (2.8~mm),
    and sensor geometry (3.84~mm width) in UE5 ensures that synthetic and real
    images share matching disparity ranges and perspective geometry, removing a domain-gap
    factor that is often overlooked in sim-to-real pipelines. Finally, UE5's \emph{Lumen
    global illumination} renders physically plausible light transport---including
    soft shadows, ambient occlusion, and leaf translucency---whose statistical
    appearance properties overlap meaningfully with those of outdoor daylight
    photography.

    \subsection{Anticipated Domain Gap}

    A gap between simulated and field-captured data will inevitably persist. The
    synthetic scenes present isolated trees against controlled backgrounds, whereas
    operational forestry imagery contains dense stands with overlapping canopies,
    undergrowth, and terrain variation; multi-tree compositions or randomised
    forest-floor textures would partially close this discrepancy. Sensor-level degradations---wind-induced
    motion blur, exposure variation, and electronic noise---are likewise absent from
    the rendered images. Additionally, the 115 Megascans models span a broad taxonomic
    range but may not capture the specific morphology of radiata pine at various
    ages and pruning stages; targeted photogrammetry of representative
    plantation specimens would improve alignment with specific deployment
    domains.

    \subsection{Relationship to Pseudo-Ground-Truth Approaches}

    Pseudo-label methods~\cite{lin2024benchmark,jiang2025defom} and synthetic rendering
    address the same data scarcity problem from opposite directions. Pseudo-labels
    operate on real imagery and therefore inherit its authentic appearance
    statistics, but they also absorb teacher-model errors---errors that are scene-dependent
    and practically impossible to validate without reference depth. Synthetic
    labels are, by construction, error-free: every disparity value is exact. The
    two strategies are complementary rather than competing: synthetic data provides
    mathematically exact supervision for pre-training, while pseudo-labels on
    real data can serve as a subsequent fine-tuning signal. UE5-Forest is designed
    to serve as the high-quality synthetic component in such hybrid pipelines.

    \subsection{Intended Use Cases}

    UE5-Forest is designed to support several research directions:
    \begin{itemize}
        \item \textbf{Supervised pre-training}: The pixel-perfect labels can serve
            as the sole or initial training signal for stereo matching networks targeting
            forestry~applications.

        \item \textbf{Domain adaptation research}: The paired availability of synthetic
            labelled data and real unlabelled Canterbury imagery provides a natural
            testbed for unsupervised domain adaptation~\cite{tonioni2019learning}
            and domain randomisation~\cite{tobin2017domain} techniques.

        \item \textbf{Benchmarking}: The held-out test split, with exact ground
            truth, enables rigorous quantitative comparison of stereo algorithms
            on vegetation~scenes.
    \end{itemize}

    \subsection{Limitations and Future Extensions}

    Several limitations bound the current dataset. The single-tree scene
    composition omits inter-tree occlusion and canopy overlap, both common in plantation
    stands. Environmental diversity is limited to UE5's default sky-light; varying
    time-of-day, cloud cover, and atmospheric scattering would improve the range
    of lighting conditions. Species coverage, while broad, does not exhaustively
    sample the morphological range of radiata pine across ages and management
    histories. Future extensions will address these gaps through multi-tree forest
    compositions, environmental-condition randomisation, and the addition of sensor-noise
    simulation to further close the domain gap.

    \section{Conclusion}

    We presented UE5-Forest, a photorealistic synthetic stereo dataset for UAV forestry
    depth estimation comprising 5,520 rectified $1920\times1080$ stereo pairs with
    pixel-perfect disparity labels. The dataset is constructed in Unreal Engine~5
    from 115 photogrammetry-scanned Quixel Megascans tree assets, with camera optics
    precisely aligned to the ZED Mini stereo rig deployed on our UAV.
    Statistical characterisation confirms that the disparity distributions are
    consistent with real-world operating conditions, and qualitative comparison with
    Canterbury Tree Branches imagery demonstrates the photorealistic quality of the
    rendered data.

    UE5-Forest fills a critical data gap: it is, to our knowledge, the first
    publicly available synthetic stereo dataset with dense, error-free disparity
    labels targeting the forestry domain. The dataset will be publicly released
    to serve as both a training resource and a benchmark for stereo matching
    research in vegetation-rich environments. Future work will extend the dataset
    with multi-tree compositions, diverse lighting conditions, and sensor-noise simulation,
    and will explore its use in hybrid training pipelines that combine synthetic
    pre-training with pseudo-label refinement on unlabelled field data.

\end{document}